\UseRawInputEncoding
\documentclass[Afour,sagev,times]{sagej}

\usepackage{moreverb,url}

\usepackage[colorlinks,bookmarksopen,bookmarksnumbered,citecolor=red,urlcolor=red]{hyperref}

\newcommand{\revision}[1]{{#1}} 

\newcommand\BibTeX{{\rmfamily B\kern-.05em \textsc{i\kern-.025em b}\kern-.08em
T\kern-.1667em\lower.7ex\hbox{E}\kern-.125emX}}

\def\volumeyear{2021}

\begin{document}

\runninghead{Strate and Christensen}

\title{Sensitivity Study of Fiducial-Aided Navigation of Unmanned Aerial Vehicles}

\author{Amanda J. Strate\affilnum{1}, Randall Christensen\affilnum{2}}

\affiliation{\affilnum{1}Graduate Research Assistant, Utah State University\\
\affilnum{2}Assistant Professor, Utah State University}

\email{randall.christensen@usu.edu}

\begin{abstract}
The possible applications and benefits of autonomous Unmanned Aerial Vehicle (UAV) use in urban areas are gaining considerable attention.  Before these possibilities can be realized, it is essential that UAVs be able to navigate reliably and precisely in urban environments.  The most common means of determining the location of a UAV is to utilize position measurements from Global Navigation Satellite Systems (GNSS).  In urban environments, however, GNSS measurements are significantly degraded due to occlusions and multipath.  This research analyzes the use of camera Line-of-Sight (LOS) measurements to self-describing fiducials as a replacement for conventional GNSS measurements.  An extended Kalman filter (EKF) is developed and validated for the purpose of combining continuous measurements from an Inertial Measurement Unit (IMU) with the discrete LOS measurements to accurately estimate the states of a UAV.  The sensitivity of the estimation error covariance to various system parameters is assessed, including IMU grade, fiducial placement, vehicle altitude, and image processing frequency.
\end{abstract}

\keywords{unmanned aerial vehicle, extended Kalman filter, inertial navigation, fiducial, GNSS-denied navigation}

\maketitle

\section{Introduction}
As the number of possible uses of UAVs in urban environments grows, so too does the importance for safe, reliable, and accurate UAV navigation.\cite{ke_real-time_2017, corning_faa_nodate}  Possible applications of UAV use in urban environments include package delivery,\cite{stolaroff_need_2014} building inspections,\cite{al-kaff_vbii-uav_2017} security,\cite{masum_simulation_2013} and emergency response coordination,\cite{rodriguez_emergency_2006} the safety requirements of which has motivated research in  UAV traffic management systems.\cite{kopardekar_unmanned_2014}

One of the challenges facing the effective use of UAVs in urban environments is that GNSS signals are often distorted or blocked entirely by surrounding infrastructure.\cite{langley_dilution_1999, el-rabbany_introduction_2002}  These signals can also be intentionally jammed.\cite{kaplan_understanding_2005}  These situations can result in position errors on the order of several meters.\cite{wing_consumer-grade_2005, modsching_field_2006}  Such large position errors are unacceptable in urban applications where the UAV must operate in close proximity to buildings and people. The need for alternative position information has motivated a large field of research, commonly referred to at GNSS-denied navigation. \revision{Two common categories of approaches to GNSS-denied navigation discussed in this paper involve using either radio-frequency (RF) or processing of optical data to determine a UAV's location within an environment}.

One RF-based approach to GNSS-denied navigation is to utilize signals of opportunity from external infrastructure \cite{kapoor_uav_2017}.  Examples include WI-FI routers,\cite{al-moukhles_multiple-fingerprints_2018, han_indoor_2017, jung_ieee_2016} cellular towers,\cite{ragothaman_multipath-optimal_2019} and television satellites.\cite{rabinowitz_position_2008} These approaches use information such as received signal strength and transmission time to estimate the position of the UAV relative to the transmitters.\cite{tseng_enhanced_2017}  Another RF-based approach is that of using pseudolites in place of GNSS satellites.\cite{huang_innovative_2019} Similar to GNSS, pseudolites are placed at surveyed locations and emit an RF signal that is used to triangulate the position of a UAV.  In all cases, the RF-based methods are susceptible to occlusion and multipath, reducing the reliability of a such a positioning system in urban environments.

An alternative \revision{to using an} RF-based method is to exploit the information contained in images captured by a camera on-board the UAV. \revision{Visual Odometry (VO) is such a method that} tracks salient features across successive images to estimate the velocity of the UAV and therefore its progressing position.\cite{koch_relative_2020, nister_visual_2004, chaolei_wang_monocular_2012}  \revision{Visual} Simultaneous Localization and Mapping (SLAM) is a common VO technique that uses multiple images of the UAV’s surroundings to simultaneously localize the observed features (i.e. generate a map) and estimate the UAV’s location within the map.\cite{chowdhary_gps-denied_2013, weiss_monocular-slam-based_2011, blosch_vision_2010, civera_inverse_2008, kim_autonomous_2004, leishman_relative_2014, wheeler_relative_2018}  Light Detection and Ranging (Lidar), which provides range in addition to intensity, can also be used in place of images to map the surroundings as the UAV progresses along its trajectory.\cite{hemann_long-range_2016}  \revision{VO approaches} provide means for relative state estimation.  Each successive estimate of the UAV’s position is dependent on the previous estimate of the UAV’s position, resulting in a gradual increase in errors over time.  \revision{In order to provide absolute estimates of position, the loop must be closed by the UAV circling back to already-traversed area.}  \revision{VO is} also very dependent on available lighting.\cite{scaramuzza_visual_2011}

This manuscript investigates a low-cost, ubiquitous alternative to GNSS: the use of self-describing fiducials (SDFs) for UAV localization.\cite{davidson_navigation_2017}  The SDFs are placed at known locations, and as a UAV traverses over the SDFs, images of the SDFs captured by a camera on-board the UAV are used to determine the UAV’s relative location to that SDF. Using fiducials for GNSS-denied navigation is not new,\cite{nguyen_remote_2017, yu_multi-resolution_2017} and multiple types of fiducials have been investigated for their reliability, data processing time, and visibility in harsh lighting conditions.\cite{claus_reliable_2004, cesar_evaluation_2015} Examples of fiducials for the purpose of SDF-aided navigation include AprilTags,\cite{olson_apriltag_2011, wang_apriltag_2016, krogius_flexible_2019} ArUco markers,\cite{cesar_evaluation_2015} ARTags,\cite{fiala_artag_2005, fiala_designing_2010} CALTags,\cite{shabalina_comparing_2017} LEDs,\cite{censi_low-latency_2013, faessler_monocular_2014} and QR codes.\cite{cavanini_qr-code_2017}  Each fiducial type comes with its own benefits and drawbacks. Some fiducials require power, while some do not. Some fiducials require daylight, while some can work in darkness. Some fiducials provide a measurement of relative position and attitude (i.e. pose), by resolving geometric features internal to the SDF, while others provide only a measurements of bearing.  \revision{The cited studies illustrate the feasibility of using fiducials for GNSS-denied navigation and discuss the strengths and weaknesses of different types of fiducials.  The expected performance and sensitivities, however, of a fiducial-aided inertial navigation system are not discussed.  This study seeks to address this gap.}

This manuscript considers the case of long distance observations of SDFs, where geometric features are not sufficiently resolved, and therefore cannot provide reliable pose estimates.  The measurement is therefore modeled as a LOS observation to a known location.  The performance sensitivities of such a fiducial-based navigation are analyzed for a relevant trajectory.  An EKF which combines continuous IMU measurements with discrete SDF measurements is developed and validated.  Covariance simulations are then used to analyze the effect of IMU grade, measurement frequency, fiducial spacing, and vehicle altitude on the position estimation error.

The remainder of this paper is organized as follows. The coordinate systems and attitude representations using in this research are discussed in the Background.  The truth model for the UAV state dynamics and measurements is discussed in the Truth Models section.  The navigation algorithms used to both propagate the states of the UAV using continuous IMU measurements as well as update those states using discrete measurements through the EKF are shown and discussed in the Navigation Algorithms section. The validation of the developed EKF is shown, and a sensitivity analysis is done in the Results section. In the final section, conclusions from this study are discussed.

\section{Background}

\begin{figure}
    \centering
    \includegraphics[width=0.45\textwidth]{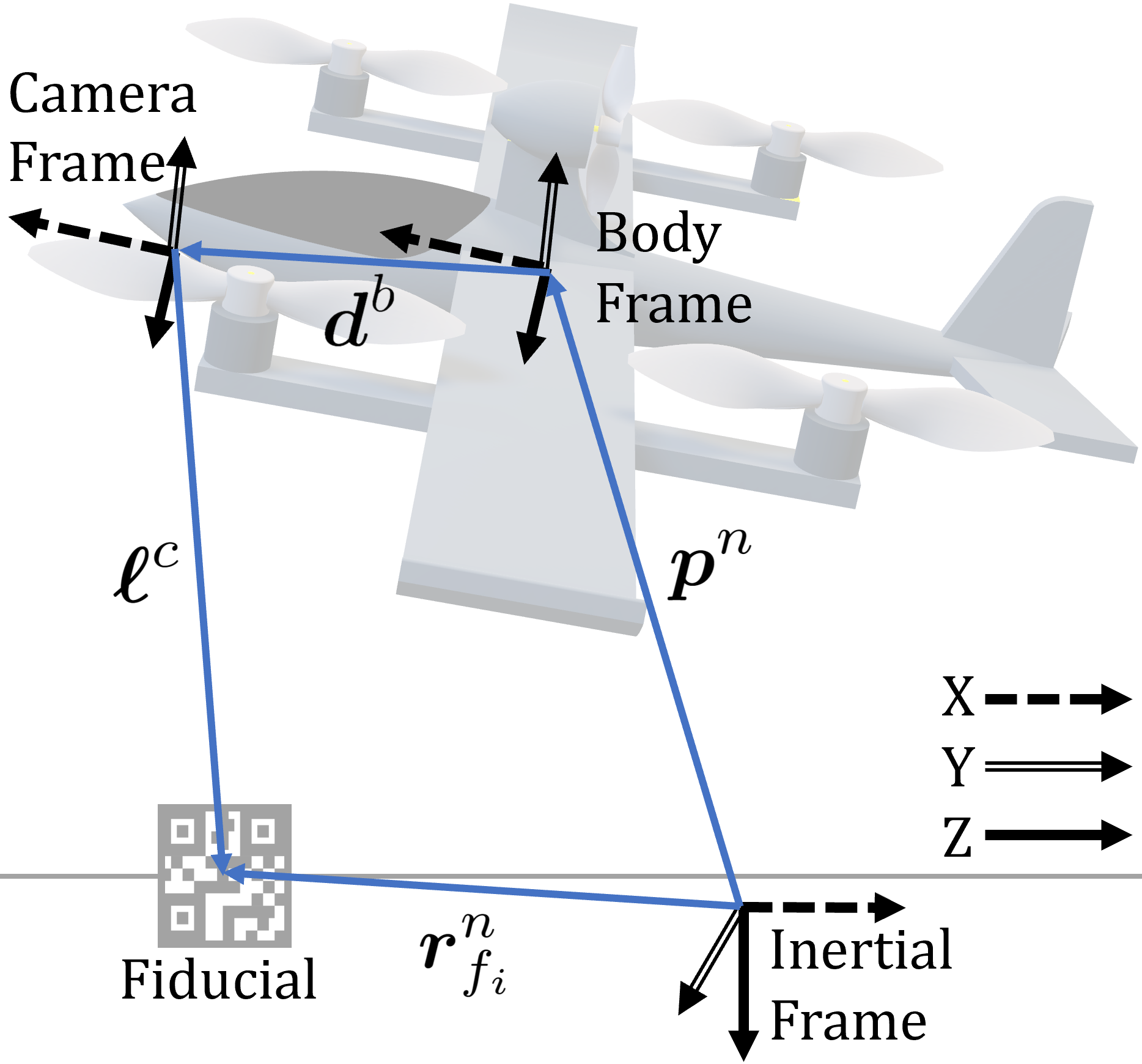}
    \caption{Inertial, body, and camera coordinate systems and measurement vectors}
    \label{fig:Coordinates}
\end{figure}

The four coordinate systems used in this study are the inertial, north-east-down (n), body (b), and camera (c) frames.  The inertial, body, and camera frames are shown in Figure \ref{fig:Coordinates}.\cite{noauthor_vtol_nodate}  The inertial frame has its origin at a fixed location on the ground and is aligned with the north-east-down frame.  The north-east-down frame has its axes pointing parallel to the axes of the inertial frame, but its origin is fixed to the navigation center of the IMU on-board the UAV.  The body frame also has its origin fixed to the navigation center of the UAV, but it rotates with the UAV. Its x-axis points out the nose of the UAV, its y-axis points down the right wing of the UAV, and the z-axis points out the belly of the UAV, following the right-hand rule.  The camera frame has its origin fixed to the focal point of the camera, and its axes are parallel to the axes of the body frame.

Many options exist for the choice of navigation framework, with varying levels of fidelity and multiple choices of coordinate systems and mechanization equations.\cite{grewal_global_2020, farrell_aided_2008, savage_strapdown_2000}  Given the short flight considered in this research, the ``Tangent Frame'' defined by Farrell\cite{farrell_aided_2008} is simplified by neglecting earth's rotation and the variations in gravity over large distances.

The quaternions used to represent attitude in this paper have the scalar portion as the first element and the vector portion as the second, third, and fourth elements.  Quaternion multiplication is performed as shown in Equation \ref{eqn:qMult} with the scalar portions of quaternions represented in non-bold font, and the vector portion is represented bolded with a subscript $v$.\cite{savage_strapdown_2000}
\begin{equation}
    \boldsymbol{q}_{1}\otimes\boldsymbol{q}_{2}=\begin{bmatrix}-\boldsymbol{q}_{1_{v}}\boldsymbol{\cdot}\boldsymbol{q}_{2_{v}}+q_{1}q_{2}\\
    \boldsymbol{q}_{1_{v}}\boldsymbol{\times}\boldsymbol{q}_{2_{v}}+q_{1}\boldsymbol{q}_{2_{v}}+q_{2}\boldsymbol{q}_{1_{v}}
    \end{bmatrix}
\label{eqn:qMult}
\end{equation}
\section{Truth models}

The states of the UAV that are estimated by the EKF in this research include the position ($\boldsymbol{p}^{n}$) and velocity ($\boldsymbol{v}^{n}$) expressed in the NED frame, attitude ($\boldsymbol{q}_b^{n}$), orientation of the camera with respect to the body frame ($\boldsymbol{q}_c^b$), and biases associated with the accelerometer ($\boldsymbol{b}_a$) and gyroscope ($\boldsymbol{b}_g$).  These states are combined into a state vector $\boldsymbol{x}$ as shown in Equation \ref{eqn:xstate}. 

\begin{equation}
    \boldsymbol{x} =  [ \quad \boldsymbol{p}^{n} \quad \boldsymbol{v}^{n} \quad \boldsymbol{q}_b^{n} \quad \boldsymbol{b}_a \quad \boldsymbol{b}_g \quad \boldsymbol{q}_c^b \quad ]^T
\label{eqn:xstate}
\end{equation}

The truth state dynamics are shown in Equation \ref{eqn:truth_de}.  The true specific force of the UAV ($\boldsymbol{\nu}^b$) is combined with gravity to calculate the velocity rate of change, the integration of which defines the rate of change in position.  The true angular velocity of the UAV ($\boldsymbol{\omega}^b$) combined with the current attitude calculates the rate of change of the attitude quaternion.  The biases associated with the accelerometer and gyroscope are both modeled as Exponentially-Correlated Random Variables  (ECRV) with a corresponding time constant ($\tau_a$ and $\tau_b$) and process noise ($\boldsymbol{n}_a$ and $\boldsymbol{n}_g$). The change in the orientation of the camera with respect to the body frame is set to 0 because the camera frame is fixed in a firm position in relation to the UAV body frame.

\begin{equation}
	\dot{\boldsymbol{x}} = 
	\begin{bmatrix}
		\dot{\boldsymbol{p}}^{n} \\[6pt]
		\dot{\boldsymbol{v}}^{n}\\[10pt]
		\dot{\boldsymbol{q}}_b^{n}\\[10pt]
		\dot{\boldsymbol{b}}_a\\[6pt]
		\dot{\boldsymbol{b}}_g\\[6pt]
		\dot{\boldsymbol{q}}_c^b
	\end{bmatrix}
	=
	\begin{bmatrix}
		\boldsymbol{v}^{n}\\[6pt]
		\boldsymbol{T}^{n}_b \, \boldsymbol{\nu}^b+\boldsymbol{g}^{n}\\[6pt]
		\frac{1}{2} \, \boldsymbol{q}_b^{n}\otimes\begin{bmatrix}0\\[6pt] \boldsymbol{\omega}^b\end{bmatrix}\\[14pt]
		-\frac{1}{\tau_a} \, \boldsymbol{b}_a+\boldsymbol{n}_a\\[6pt]
		-\frac{1}{\tau_g} \, \boldsymbol{b}_g+\boldsymbol{n}_g\\[6pt]
	    \boldsymbol{0}
	\end{bmatrix}		
\label{eqn:truth_de}
\end{equation}

The IMU provides continuous measurements of specific force and angular rate, corrupted by bias and noise.
\begin{equation}
	\tilde{\boldsymbol{\nu}}^b=\boldsymbol{\nu}^b+\boldsymbol{b}_{a}+\boldsymbol{n}_{\nu}
\label{eqn:nutildedef}
\end{equation}
\begin{equation}
\tilde{	\boldsymbol{\omega}}^b=\boldsymbol{\omega}^b+\boldsymbol{b}_{g}+\boldsymbol{n}_{\omega}
\label{eqn:omegatildedef}
\end{equation}
In both cases, the noise is modeled as zero-mean, white Gaussian noise, with power spectral density defined as
\begin{equation}
    E\left[\boldsymbol{n_{\nu}}(t)[\boldsymbol{n_{\nu}}(t')^T\right]=q_{\nu}\boldsymbol{I}_{3\times3}\delta(t-t')
\end{equation}
\begin{equation}
    E\left[\boldsymbol{n_{\omega}}(t)[\boldsymbol{n_{\omega}}(t')^T\right]=q_{\omega}\boldsymbol{I}_{3\times3}\delta(t-t')
\end{equation}
where $\sqrt{q_{\nu}}$ and $\sqrt{q_{\omega}}$ are commonly referred to as the \emph{velocity random walk} (VRW) and \emph{angular random walk} (ARW) specifications, respectively.

The discrete measurements available to the EKF consist of position measurements from a loosely-coupled GNSS and LOS measurements from the on-board camera.

A GNSS position measurement is used during the initial portions of the simulation when the UAV is at an altitude that is considered to be above buildings and therefore in an area where GNSS signals are valid and fiducial measurements may not be available.  The GNSS measurement is simply the true position of the UAV corrupted by zero-mean, Gaussian noise $\boldsymbol{n}_{\text{GNSS}}$ 
\begin{equation}
    \tilde{\boldsymbol{z}}_{\text{GNSS}}=\boldsymbol{p}^{n}+\boldsymbol{n}_{\text{GNSS}}
\label{eqn:gnssMeas}
\end{equation}
where the correlation kernel of $\boldsymbol{n}_{\text{GNSS}}$ is defined as
\begin{equation}
    \boldsymbol{E}\left[\boldsymbol{n}_{GNSS}\left(t_i\right)\boldsymbol{n}_{\text{GNSS}}\left(t_j\right)\right] = \boldsymbol{\sigma}_{\text{GNSS}}\boldsymbol{I}_{3\times3}\delta\left[t_i,t_j\right]
\end{equation}

The LOS vector from the camera to the fiducial is illustrated in Figure \ref{fig:Coordinates} and is a combination of the UAV position, the known position of the fiducial $\boldsymbol{r}_{f_i}^{n}$, and the known displacement from the UAV body frame to the camera frame $\boldsymbol{d}^b$.
\begin{equation}
	\boldsymbol{\ell}^c
	=
	\boldsymbol{T}_b^c \left[\boldsymbol{T}_{n}^b \left(\boldsymbol{r}_{f_i}^{n} - \boldsymbol{p}^{n}\right) - \boldsymbol{d}^b\right]
\label{eqn:losc}
\end{equation}

The measurement obtained from the camera is the projection of the LOS vector to the image plane, following a pin-hole camera model, and it is corrupted by zero-mean, Gaussian noise $\boldsymbol{n}_\ell$
\begin{equation}
    \tilde{\boldsymbol{z}}_{\text{LOS}}=\begin{bmatrix}\frac{\ell_{x}^{c}}{\ell_{z}^{c}} \\[6pt] \frac{\ell_{y}^{c}} {\ell_{z}^{c}}\end{bmatrix} + \boldsymbol{n}_{\ell}
    \label{eqn:zTildeLOSTruth}
\end{equation}
where the correlation kernel of $\boldsymbol{n}_\ell$ is defined as
\begin{equation}
    \boldsymbol{E}\left[\boldsymbol{n}_\ell\left(t_i\right)\boldsymbol{n}_\ell\left(t_j\right)\right] = \boldsymbol{\sigma}_{\ell} \boldsymbol{I}_{2\times2}\delta\left[t_i,t_j\right]
\end{equation}
\section{Navigation algorithms}

The models used for the navigation states are consistent with the models used for the truth state, with a few subtle differences.  Equation \ref{eqn:nav_de} shows the differential equation used to propagate the navigation (or estimated) states of the UAV. Note that the estimated biases for the accelerometer and gyroscope are subtracted from the accelerometer and gyroscope measurements to obtain an estimate of the true specific force and angular rate.  Furthermore, the ECRV's are propagated in the absence of the associated process noise since such quantities are unknown to the EKF.

\begin{equation}
	\dot{\hat{\boldsymbol{x}}} 
	= 
	\begin{bmatrix}
		\dot{\hat{\boldsymbol{p}}}^{n}\\[8pt]
		\dot{\hat{\boldsymbol{v}}}^{n}\\[12pt]
		\dot{\hat{\boldsymbol{q}}}_{b}^{n}\\[12pt]
		\dot{\hat{\boldsymbol{b}}}_a\\[6pt]
		\dot{\hat{\boldsymbol{b}}}_g\\[6pt]
		\dot{\hat{\boldsymbol{q}}}_{c}^b
	\end{bmatrix}
	=
	\begin{bmatrix}
		\hat{\boldsymbol{v}}^{n}\\[6pt]
		\hat{\boldsymbol{T}}_{b}^{n} \, (\tilde{\boldsymbol{\nu}}^b-\hat{\boldsymbol{b}}_a)+\boldsymbol{g}^{n}\\[6pt]
		\frac{1}{2} \,
	    \boldsymbol{q}_{\hat{b}}^{n} \otimes
		\begin{bmatrix}0\\[6pt]
		    \tilde{\boldsymbol{\omega}}^b-\hat{\boldsymbol{b}}_g
	    \end{bmatrix}\\[12pt]
	    -\frac{1}{\tau_a} \, \hat{\boldsymbol{b}}_a\\[6pt]
	    -\frac{1}{\tau_g} \, \hat{\boldsymbol{b}}_g\\[6pt]
	    \boldsymbol{0}
	\end{bmatrix}	
\label{eqn:nav_de}
\end{equation}

To facilitate the propagation of the covariance matrix $P$, Equation \ref{eqn:truth_de} is linearized about the current state estimate to obtain
\begin{equation}
    \delta\dot{\boldsymbol{x}}=\boldsymbol{F}\left(\hat{\boldsymbol{x}}\right)\delta\boldsymbol{x}+\boldsymbol{B}\left(\hat{\boldsymbol{x}}\right)\boldsymbol{w}
\label{eqn:linearErrorGen}
\end{equation}
where the state dynamics matrix $F$ is defined by the following equations.
\begin{equation}
    \boldsymbol{F}
    =
    \begin{bmatrix}
        \boldsymbol{0}_{3\times3} & \boldsymbol{I}_{3\times3} & \boldsymbol{0}_{3\times3} & \boldsymbol{0}_{3\times3} & \boldsymbol{0}_{3\times3} & \boldsymbol{0}_{3\times3}\\
        \boldsymbol{0}_{3\times3} & \boldsymbol{0}_{3\times3} & \boldsymbol{F}_{\text{att}} & -\hat{\boldsymbol{T}}_{b}^{n} & \boldsymbol{0}_{3\times3} & \boldsymbol{0}_{3\times3}\\
        \boldsymbol{0}_{3\times3} & \boldsymbol{0}_{3\times3} & \boldsymbol{0}_{3\times3} & \boldsymbol{0}_{3\times3} & \hat{\boldsymbol{T}}_{b}^{n} & \boldsymbol{0}_{3\times3}\\
        \boldsymbol{0}_{3\times3} & \boldsymbol{0}_{3\times3} & \boldsymbol{0}_{3\times3} & \boldsymbol{F}_{\text{accel}} & \boldsymbol{0}_{3\times3} & \boldsymbol{0}_{3\times3}\\
        \boldsymbol{0}_{3\times3} & \boldsymbol{0}_{3\times3} & \boldsymbol{0}_{3\times3} & \boldsymbol{0}_{3\times3} & \boldsymbol{F}_{\text{gyro}} & \boldsymbol{0}_{3\times3}\\
        \boldsymbol{0}_{3\times3} & \boldsymbol{0}_{3\times3} & \boldsymbol{0}_{3\times3} & \boldsymbol{0}_{3\times3} & \boldsymbol{0}_{3\times3} & \boldsymbol{0}_{3\times3}
    \end{bmatrix}
\label{eqn:LinF}
\end{equation}
\begin{equation}
    \boldsymbol{F}_{\text{att}} =  \left(\left(\hat{\boldsymbol{T}}_{b}^{n}\left(\tilde{\boldsymbol{\nu}}^{b}-\hat{\boldsymbol{b}}_{a}\right)\right)\times\right)
\end{equation}
\begin{equation}
    \boldsymbol{F}_{\text{accel}} = -\frac{1}{\tau_{a}}\boldsymbol{I}_{3\times3}
\end{equation}
\begin{equation}
    \boldsymbol{F}_{\text{gyro}} = -\frac{1}{\tau_{g}}\boldsymbol{I}_{3\times3}
    \label{eqn:FGyro}
\end{equation}

The noise coupling matrix $B$ and associated process noise vector $\boldsymbol{w}$ are define as follows.
\begin{equation}
    \boldsymbol{B}
    =
    \begin{bmatrix}
        \boldsymbol{0}_{3\times3} & \boldsymbol{0}_{3\times3} & \boldsymbol{0}_{3\times3} & \boldsymbol{0}_{3\times3}\\
        -\hat{\boldsymbol{T}}_{b}^{n} & \boldsymbol{0}_{3\times3} & \boldsymbol{0}_{3\times3} & \boldsymbol{0}_{3\times3}\\
        \boldsymbol{0}_{3\times3} & \hat{\boldsymbol{T}}_{b}^{n} & \boldsymbol{0}_{3\times3} & \boldsymbol{0}_{3\times3}\\
        \boldsymbol{0}_{3\times3} & \boldsymbol{0}_{3\times3} & \boldsymbol{I}_{3\times3} & \boldsymbol{0}_{3\times3}\\
        \boldsymbol{0}_{3\times3} & \boldsymbol{0}_{3\times3} & \boldsymbol{0}_{3\times3} & \boldsymbol{I}_{3\times3}\\
        \boldsymbol{0}_{3\times3} & \boldsymbol{0}_{3\times3} & \boldsymbol{0}_{3\times3} & \boldsymbol{0}_{3\times3}
    \end{bmatrix}
\label{eqn:LinB}
\end{equation}
\begin{equation}
    \boldsymbol{w}
    =
    \begin{bmatrix}\boldsymbol{n}_{\nu}\\
        \boldsymbol{n}_{\omega}\\
        \boldsymbol{n}_{a}\\
        \boldsymbol{n}_{g}
    \end{bmatrix}
\label{eqn:noiseW}
\end{equation}

The estimation error covariance matrix $P$ is propagated according to the continuous Ricatti equation
\begin{equation}
    \dot{\boldsymbol{P}}=\boldsymbol{F}\boldsymbol{P}+\boldsymbol{P}\boldsymbol{F}^{T}+\boldsymbol{B}\boldsymbol{Q}\boldsymbol{B}^{T}
\label{eqn:CovProp}
\end{equation}
where the power spectral density matrix $Q$ is defined as
\begin{equation}
    \boldsymbol{Q}
    =
    \begin{bmatrix}
        \boldsymbol{Q}_{\boldsymbol{n}_{\nu}} & \boldsymbol{0}_{3\times3} & \boldsymbol{0}_{3\times3} & \boldsymbol{0}_{3\times3}\\
        \boldsymbol{0}_{3\times3} & \boldsymbol{Q}_{\boldsymbol{n}_{\omega}} & \boldsymbol{0}_{3\times3} & \boldsymbol{0}_{3\times3}\\
        \boldsymbol{0}_{3\times3} & \boldsymbol{0}_{3\times3} & \boldsymbol{Q}_{\boldsymbol{n}_a} & \boldsymbol{0}_{3\times3}\\
        \boldsymbol{0}_{3\times3} & \boldsymbol{0}_{3\times3} & \boldsymbol{0}_{3\times3} & \boldsymbol{Q}_{\boldsymbol{n}_g}
    \end{bmatrix}
\label{eqn:NoiseStrength}
\end{equation}
For the accelerometer and gyroscope noise, the power spectral density matrices are related to the VRW and ARW specifications.
\begin{equation}
    \boldsymbol{Q}_{\boldsymbol{n}_{\nu}}
    = q_{\nu} \boldsymbol{I}_{3\times3}
\label{eqn:Qetaa}
\end{equation}
\begin{equation}
    \boldsymbol{Q}_{\boldsymbol{n}_{\omega}}
    = q_{\omega} \boldsymbol{I}_{3\times3}
\label{eqn:Qetag}
\end{equation}
For the biases, however, the power spectral density is related to the corresponding steady-state standard deviation $\sigma_{ss}$ and time constant $\tau$.
\begin{equation}
    \boldsymbol{Q}_{\boldsymbol{n}_a}
    =
    \frac{2\sigma_{\text{ss},a}^{2}}{\tau_{a}}\boldsymbol{I}_{3\times3}
\label{eqn:Qa}
\end{equation}
\begin{equation}
    \boldsymbol{Q}_{\boldsymbol{n}_g}=\frac{2\sigma_{\text{ss},g}^{2}}{\tau_{g}}\boldsymbol{I}_{3\times3}
\label{eqn:Qg}
\end{equation}

Upon reception of either a GNSS or LOS measurement, the posterior estimate of the error state vector is calculated as
\begin{equation}
    \delta\boldsymbol{x}^+ = \boldsymbol{K}\left(\tilde{\boldsymbol{z}}-\hat{\tilde{\boldsymbol{z}}}\right)
\label{eqn:delx}
\end{equation}
where $\tilde{\boldsymbol{z}}$ is the measurement and $\hat{\tilde{\boldsymbol{z}}}$ is the predicted measurement.  The posterior covariance is calculated as
\begin{align}
    \boldsymbol{P}^{+}
    =
    \left[\boldsymbol{I}-\boldsymbol{K}\boldsymbol{H}\left(\hat{\boldsymbol{x}}^{-}\right)\right]\boldsymbol{P}^{-}\left[\boldsymbol{I}-\boldsymbol{K}\boldsymbol{H}\left(\hat{\boldsymbol{x}}^{-}\right)\right]^{T}+\notag\\
    \boldsymbol{K}\boldsymbol{G}\boldsymbol{R}\boldsymbol{G}^{T}\boldsymbol{K}^{T}
\label{eqn:EstCapCov}
\end{align}
The Kalman gain $\boldsymbol{K}$ is calculated using the apriori covariance matrix $\boldsymbol{P}^{-}$ and as well as the sensitivity matrix $\boldsymbol{H}\left(\hat{\boldsymbol{x}}^{-}\right)$, covariance matrix $\boldsymbol{R}$, and noise coupling matrix $\boldsymbol{G}$ for the available measurement.
\begin{equation}
    \boldsymbol{K}
    =
    \boldsymbol{P}^{-}\boldsymbol{H}^{T}\left(\hat{\boldsymbol{x}}^{-}\right) \boldsymbol{P}_{\text{residual}}^{-1}
\label{eqn:EstCapKalmanGain}
\end{equation}
where
\begin{equation}
    \boldsymbol{P}_{\text{residual}} = \boldsymbol{H}\left(\hat{\boldsymbol{x}}^{-}\right)\boldsymbol{P}^{-}\boldsymbol{H}^{T}\left(\hat{\boldsymbol{x}}^{-}\right)+\boldsymbol{G}\boldsymbol{R}\boldsymbol{G}^{T}
\end{equation}

The predicted LOS measurement is calculated as shown in Equations \ref{eqn:ellhat} and \ref{eqn:zTildeLOS}.
\begin{equation}
    \hat{\boldsymbol{\ell}}^{c} = \boldsymbol{T}_b^c \left[\boldsymbol{T}_{n}^b \left(\boldsymbol{r}_{f_i}^{n} - \hat{\boldsymbol{p}}^{n}\right) - \boldsymbol{d}^b\right]
    \label{eqn:ellhat}
\end{equation}
\begin{equation}
    \hat{\tilde{\boldsymbol{z}}}_{\text{LOS}} = \begin{bmatrix}\frac{\hat{\ell}_{x}^{c}}{\hat{\ell}_{z}^{c}} \\[6pt] \frac{\hat{\ell}_{y}^{c}} {\hat{\ell}_{z}^{c}}\end{bmatrix}
    \label{eqn:zTildeLOS}
\end{equation}

The predicted GNSS measurement is calculated as shown in Equation \ref{eqn:zTildeGNSS}.
\begin{equation}
    \hat{\tilde{\boldsymbol{z}}}_{\text{GNSS}} = \hat{\boldsymbol{p}}^{n}
    \label{eqn:zTildeGNSS}
\end{equation}

The GNSS and LOS measurements from Equations \ref{eqn:gnssMeas} and \ref{eqn:zTildeLOSTruth} are linearized to be in the form shown in Equation \ref{eqn:zLin}, which yields the sensitivity matrix $\boldsymbol{H}$ and the noise coupling matrix $\boldsymbol{G}$.  

\begin{equation}
    \delta\tilde{\boldsymbol{z}} = \boldsymbol{H}\left(\hat{\boldsymbol{x}}\right)\delta\boldsymbol{x}+\boldsymbol{G}\boldsymbol{n}
    \label{eqn:zLin}
\end{equation}

For the LOS measurements, the measurement covariance matrix is as shown in Equation \ref{eqn:RLOS}, the sensitivity matrix is as shown in Equations \ref{eqn:HLOSComb} through \ref{eqn:losMeasLinThetaC}, and the noise coupling matrix is as shown in Equation \ref{eqn:GLOS}.
\begin{equation}
    \boldsymbol{R}
    =
    \begin{bmatrix}
        \sigma_{\ell_{x}}^{2} & 0\\
        0 & \sigma_{\ell_{y}}^{2}
    \end{bmatrix}
\label{eqn:RLOS}
\end{equation}
\begin{equation}
    \boldsymbol{H}_{\text{LOS}}
    =
    \boldsymbol{H}_{\boldsymbol{\ell}}\boldsymbol{H}_{\text{C}}
    \label{eqn:HLOSComb}
\end{equation}
\begin{equation}
    \boldsymbol{H}_{\boldsymbol{\ell}} =
    \begin{bmatrix}\frac{1}{\hat{\ell}_{z}} & 0 & -\frac{\hat{\ell}_{x}}{\hat{\ell}_{z}^{2}}\\
    0 & \frac{1}{\hat{\ell}_{z}} & -\frac{\hat{\ell}_{y}}{\hat{\ell}_{z}^{2}}
    \end{bmatrix}
    \label{eqn:H_ell}
\end{equation}
\begin{equation}
    \boldsymbol{H}_{\text{C}}=\begin{bmatrix}\boldsymbol{H}_{\delta\boldsymbol{p}^{n}} & \boldsymbol{0}_{3\times3} & \boldsymbol{H}_{\delta\boldsymbol{\theta}_{b}^{n}} & \boldsymbol{0}_{3\times6} & \boldsymbol{H}_{\delta\boldsymbol{\theta}_{c}^{b}}\end{bmatrix}
    \label{eqn:H_LOS}
\end{equation}
\begin{equation}
    H_{\delta \boldsymbol{p}^{n}}
    =
    -\hat{\boldsymbol{R}}_{b}^{c}\hat{\boldsymbol{R}}_{n}^{b}
\label{eqn:losMeasLinP}
\end{equation}
\begin{equation}
    \boldsymbol{H}_{\delta\boldsymbol{\theta}_{b}^{n}}
    =
    \hat{\boldsymbol{R}}_{b}^{c}\hat{\boldsymbol{R}}_{n}^{b}\left[\left(\hat{\boldsymbol{p}}^{n}-\boldsymbol{r}_{f_{i}}^{n}\right)\times\right]
\label{eqn:losMeasLinThetaB}
\end{equation}
\begin{equation}
    \boldsymbol{H}_{\delta\boldsymbol{\theta}_{c}^{b}}
    =
    -\hat{\boldsymbol{R}}_{b}^{c}\left\{ \left[\hat{\boldsymbol{R}}_{n}^{b}\left(\boldsymbol{r}_{f_{i}}^{n}-\hat{\boldsymbol{p}}^{n}\right)-\boldsymbol{d}^b\right]\times\right\}
\label{eqn:losMeasLinThetaC}
\end{equation}
\begin{equation}
    \boldsymbol{G}
    = \boldsymbol{I}_{2\times2}
    \label{eqn:GLOS}
\end{equation}

For the GNSS measurements, the covariance matrix is as shown in Equation \ref{eqn:RGNSS}, the sensitivity matrix is as shown in Equation \ref{eqn:HGNSS}, and the noise coupling matrix is as shown in Equation \ref{eqn:GGNSS}.
\begin{equation}
    \boldsymbol{R}=\begin{bmatrix}\sigma_{\text{GNSS}_{x}} & 0 & 0\\
    0 & \sigma_{\text{GNSS}_{y}} & 0\\
    0 & 0 & \sigma_{\text{GNSS}_{z}}
    \end{bmatrix}
    \label{eqn:RGNSS}
\end{equation}
\begin{equation}
    \boldsymbol{H}=\begin{bmatrix}\boldsymbol{I}_{3\times3} & \boldsymbol{0}_{3\times15}\end{bmatrix}
    \label{eqn:HGNSS}
\end{equation}
\begin{equation}
    \boldsymbol{G}=\boldsymbol{I}_{3\times3}
    \label{eqn:GGNSS}
\end{equation}

Finally, the estimated error state vector from Equation \ref{eqn:delx} is used to correct the estimated states of the UAV.  
\begin{equation}
	\hat{\boldsymbol{x}}_{\text{updated}}
	=
	\begin{bmatrix}
		\hat{\boldsymbol{p}}^{n} + \delta\boldsymbol{p}^{n}\\[6pt]
		\hat{\boldsymbol{v}}^{n} + \delta\boldsymbol{v}^{n}\\[6pt]
		\begin{bmatrix}
		    1\\[6pt]
		    -\frac{1}{2}\delta\boldsymbol{\theta}_b^{n}
		\end{bmatrix} \otimes \hat{q}_{b}^{n}\\[6pt]
		\hat{\boldsymbol{b}}_a+\delta\boldsymbol{b}_a\\[6pt]
		\hat{\boldsymbol{b}}_g+\delta\boldsymbol{b}_g\\[6pt]
		\begin{bmatrix}
		    1\\[6pt]
		    -\frac{1}{2}\delta\boldsymbol{\theta}_c^b
		\end{bmatrix} \otimes \hat{q}_{c}^b\\[6pt]
	\end{bmatrix}
	\label{eqn:xupdated}
\end{equation}
\section{Results}

\begin{figure}
    \centering
    \includegraphics[trim={0.5cm 4.9cm 1.58cm 5.7cm},clip,width=0.45\textwidth]{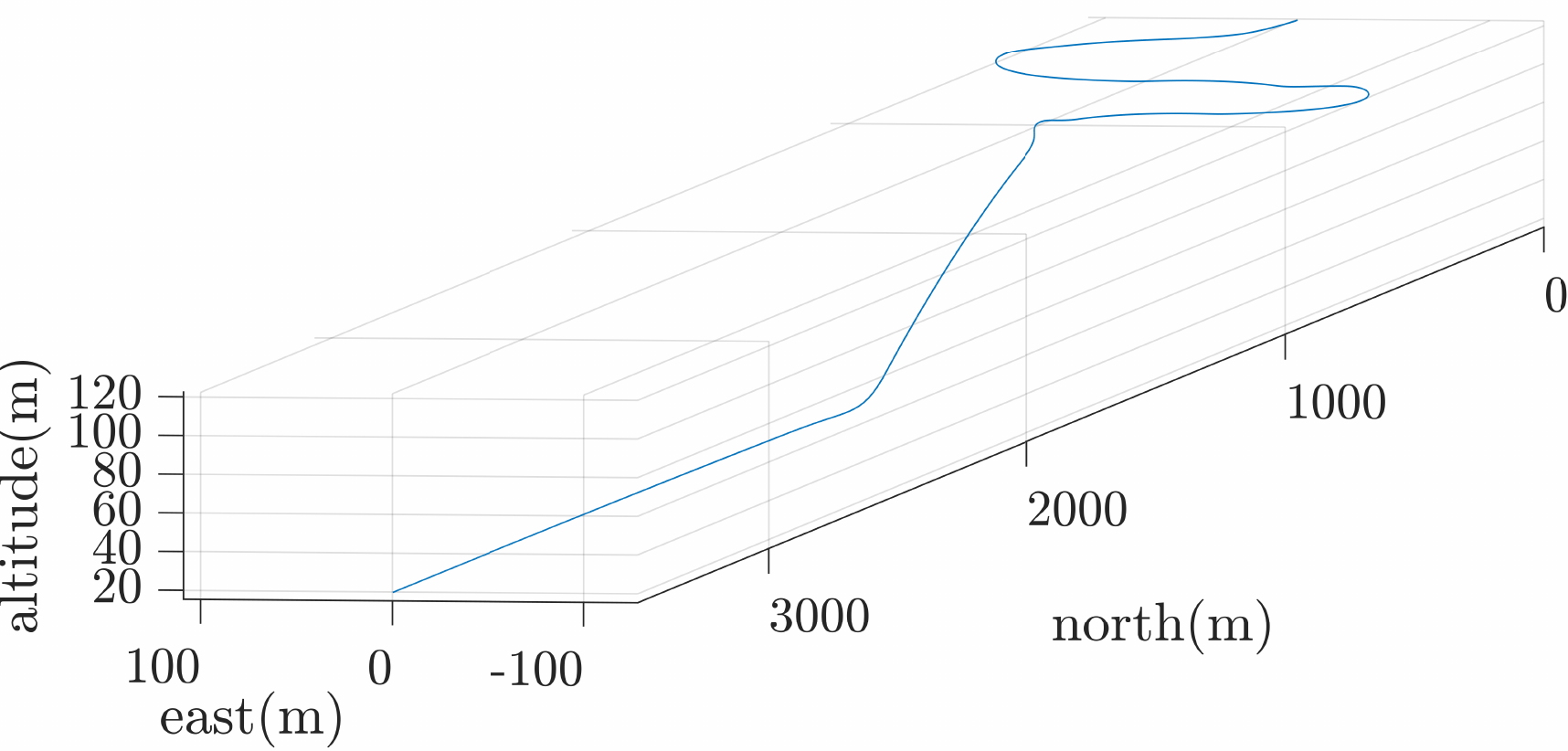}
    \caption{Simulated trajectory}
    \label{fig:Trajectory}
\end{figure}

The simulated UAV flight trajectory is shown in Figure \ref{fig:Trajectory}.  The trajectory begins with S-curves at an altitude of 120 meters where GNSS signals are available. After the S-curves, the UAV descends below an altitude of 40 meters, which is the point at which GNSS signals are considered to be lost or no longer reliable, and the UAV begins to rely on LOS measurements to fiducials.  This happens about 48.5 seconds into the simulation.  The entire simulation lasts for about 105 seconds.  During the LOS processing portion of the flight, the fiducials are placed on alternating sides of the UAV's ground trace 6 meters to either side.  The nominal height of the trajectory above the fiducials was set to 15 meters.  The nominal LOS measurement frequency is set to 5 Hz, and the nominal north distance between fiducials is set to 100 meters.  The nominal IMU grade corresponds to a tactical IMU.  GNSS measurements are processed at a frequency of 1 Hz.  A 40 degree field of view (FOV) is used for the camera, and only fiducials that are within this field of view are processed. If multiple fiducials are visible in a single image, each fiducial is processed as a separate measurement at the same time step.

The specifications for commercial-, tactical-, and navigation-grade IMU are shown in Table \ref{table:IMUGrades}.

\begin{table}
\small\sf\centering
\caption{IMU specifications by grade}
\label{table:IMUGrades}
\begin{tabular}{llll}
\toprule
 & Commercial & Tactical & Navigation \\
\midrule
$\sigma_{\text{ss},a}\ (\text{g})$ & 0.0100 & 0.0010 & 0.0001\\
$q_\nu$ $(\frac{\text{m}}{\text{s}}/\sqrt{\text{hr}})$ & 0.600 & 0.060 & 0.006\\
$\sigma_{\text{ss},g}\ (^{\circ}/\text{hr})$ & 10.0 & 1.0 & 0.1\\
$q_\omega$ $(^{\circ}/\sqrt{\text{hr}})$ & 0.700 & 0.070 & 0.007\\
\bottomrule
\end{tabular}
\end{table}

The statistics for the noise associated with the LOS and GNSS measurements is shown in Table \ref{table:MeasUncertainty}.


\begin{table}
\small\sf\centering
\caption{Measurement uncertainty}
\label{table:MeasUncertainty}
\begin{tabular}{ll}
\toprule
Statistic & 3-$\sigma$ Value\\\midrule
3-$\sigma_{\text{GNSS}_x}$  & 1.00 meter\\
3-$\sigma_{\text{GNSS}_y}$ & 1.00 meter\\
3-$\sigma_{\text{GNSS}_z}$ & 3.00 meters\\
3-$\sigma_{\ell_x}$ & 3.64 milliradians\\
3-$\sigma_{\ell_y}$ & 3.64 milliradians\\
\bottomrule
\end{tabular}
\end{table}

\subsection{Validation}

\begin{figure}
    \centering
    \includegraphics[trim={0.7cm 0.2cm 1.4cm 0.4cm},clip,width=0.45\textwidth]{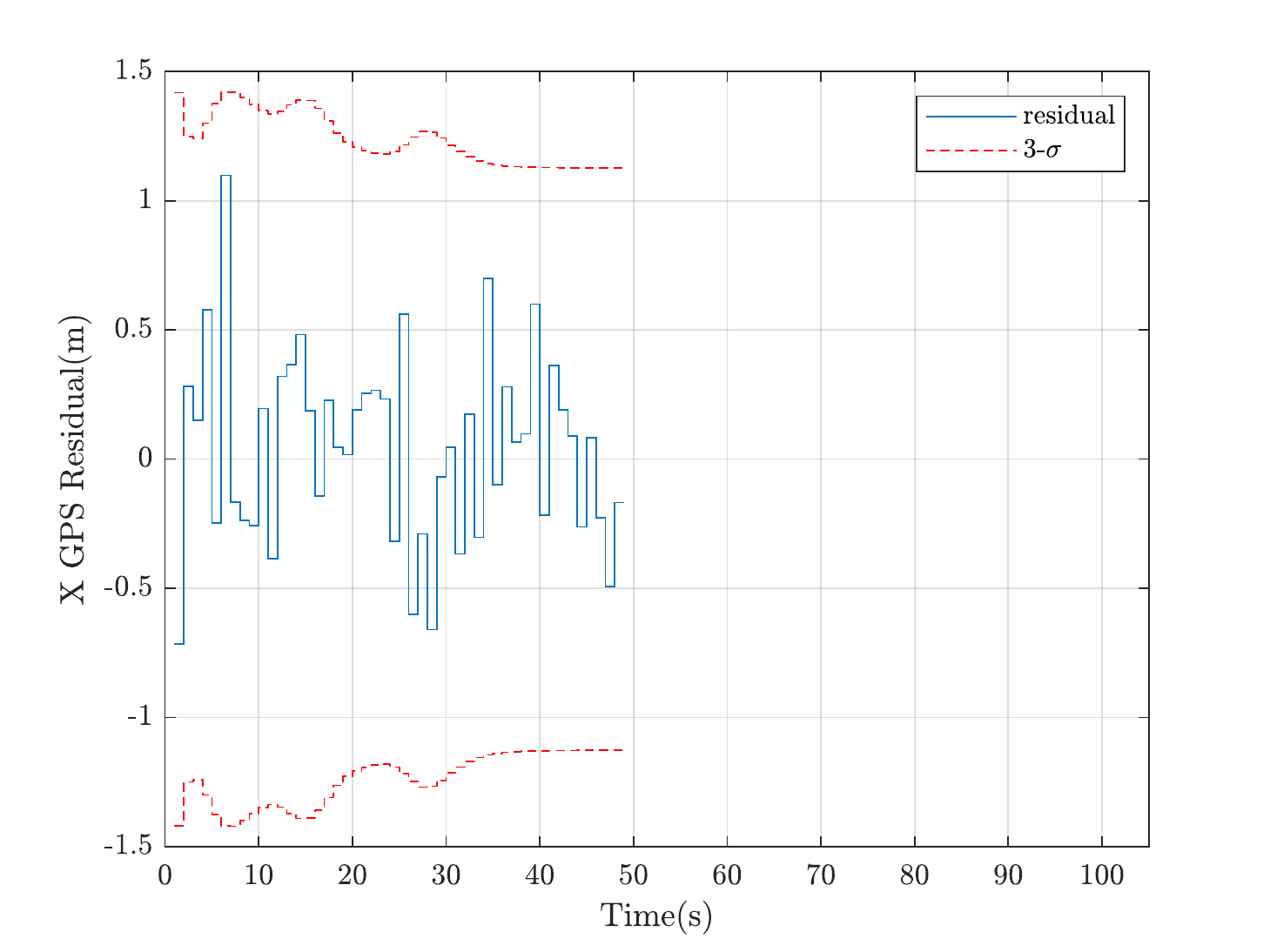}
    \includegraphics[trim={0.7cm 0.2cm 1.4cm 0.4cm},clip,width=0.45\textwidth]{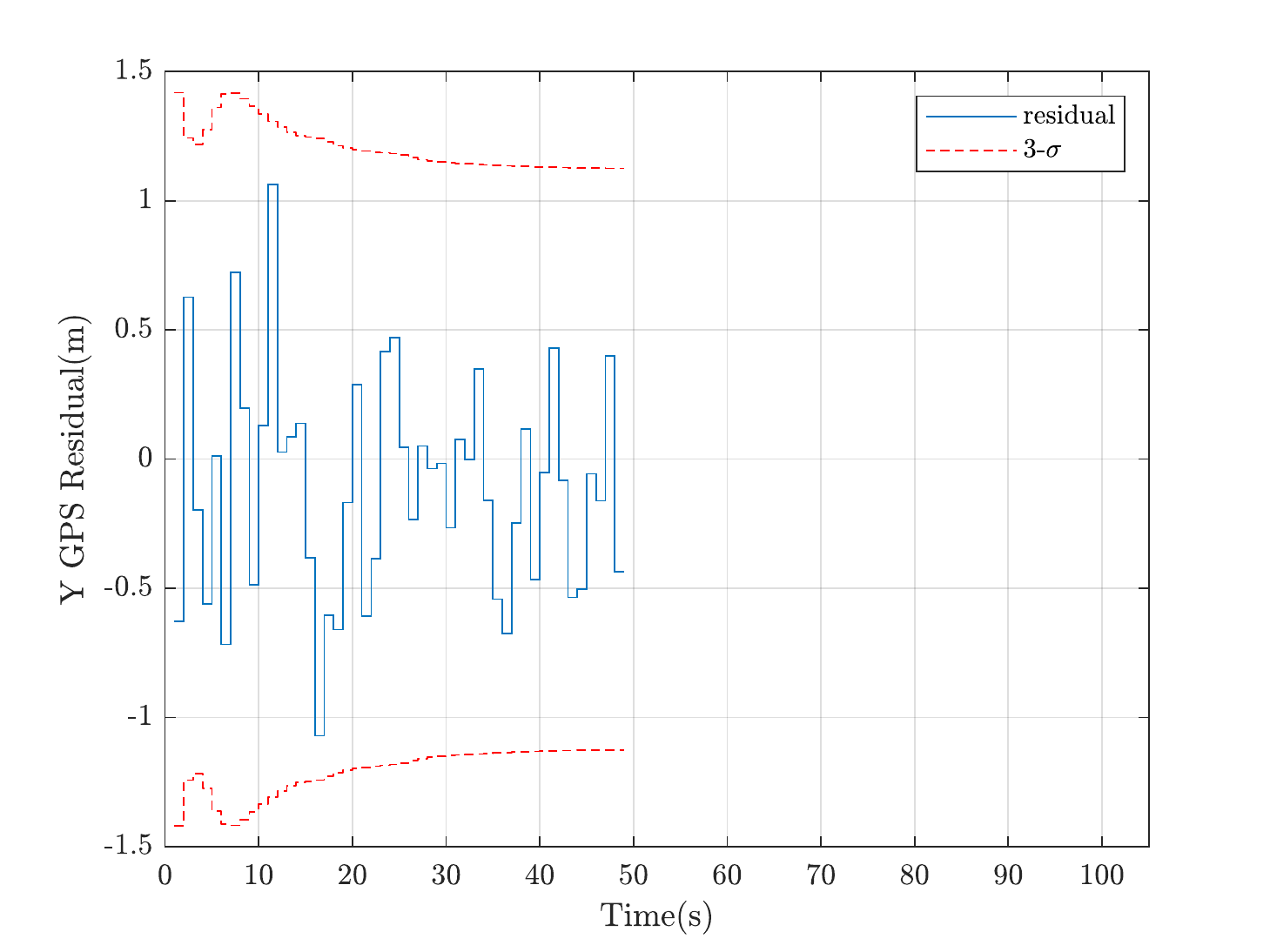}
    \includegraphics[trim={0.7cm 0.2cm 1.4cm 0.4cm},clip,width=0.45\textwidth]{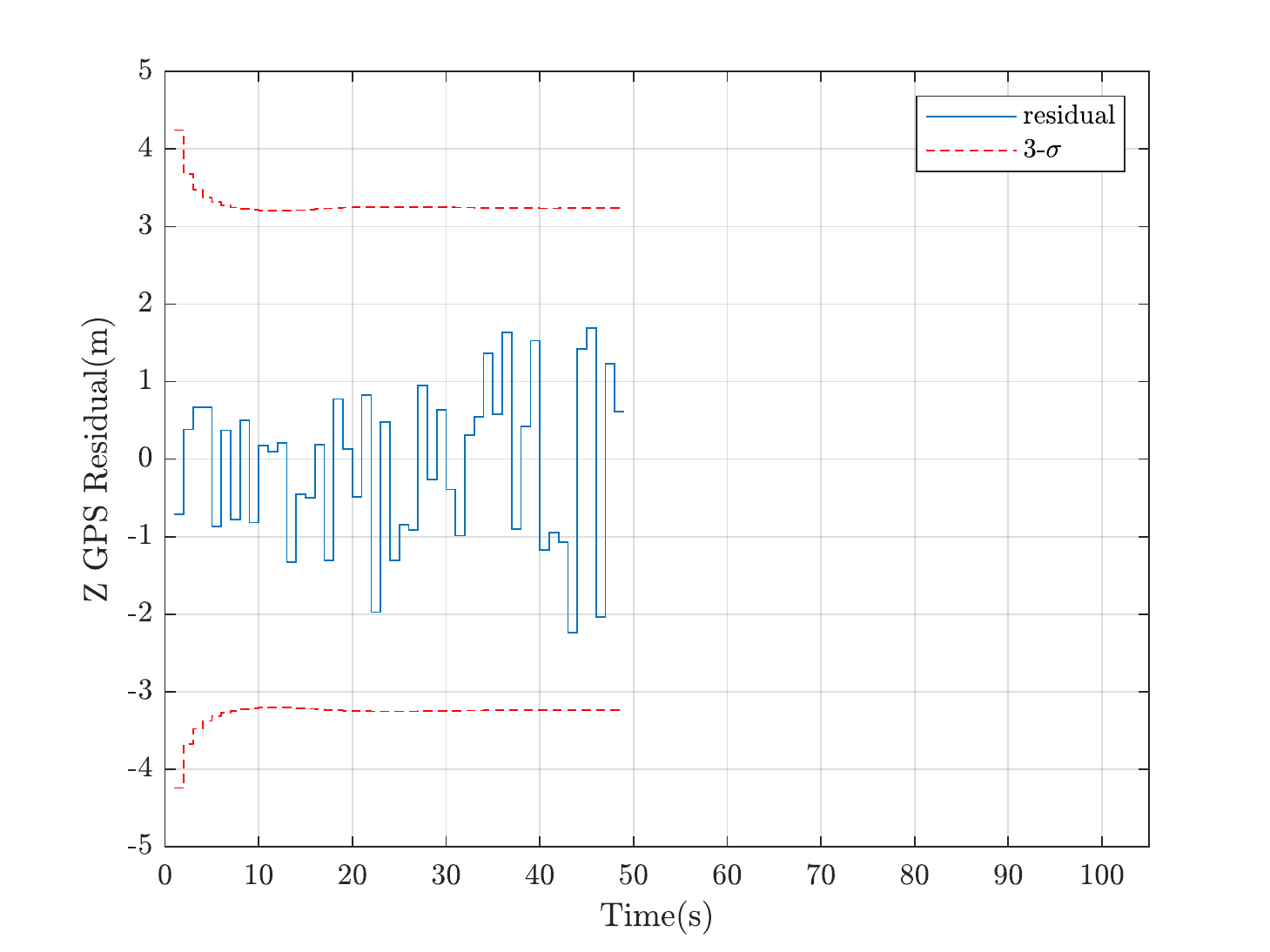}
    \caption{GNSS measurement residuals}
    \label{fig:GNSSResiduals}
\end{figure}

\begin{figure}
    \centering
    \includegraphics[trim={0.9cm 0.2cm 1.4cm 0.4cm},clip,width=0.45\textwidth]{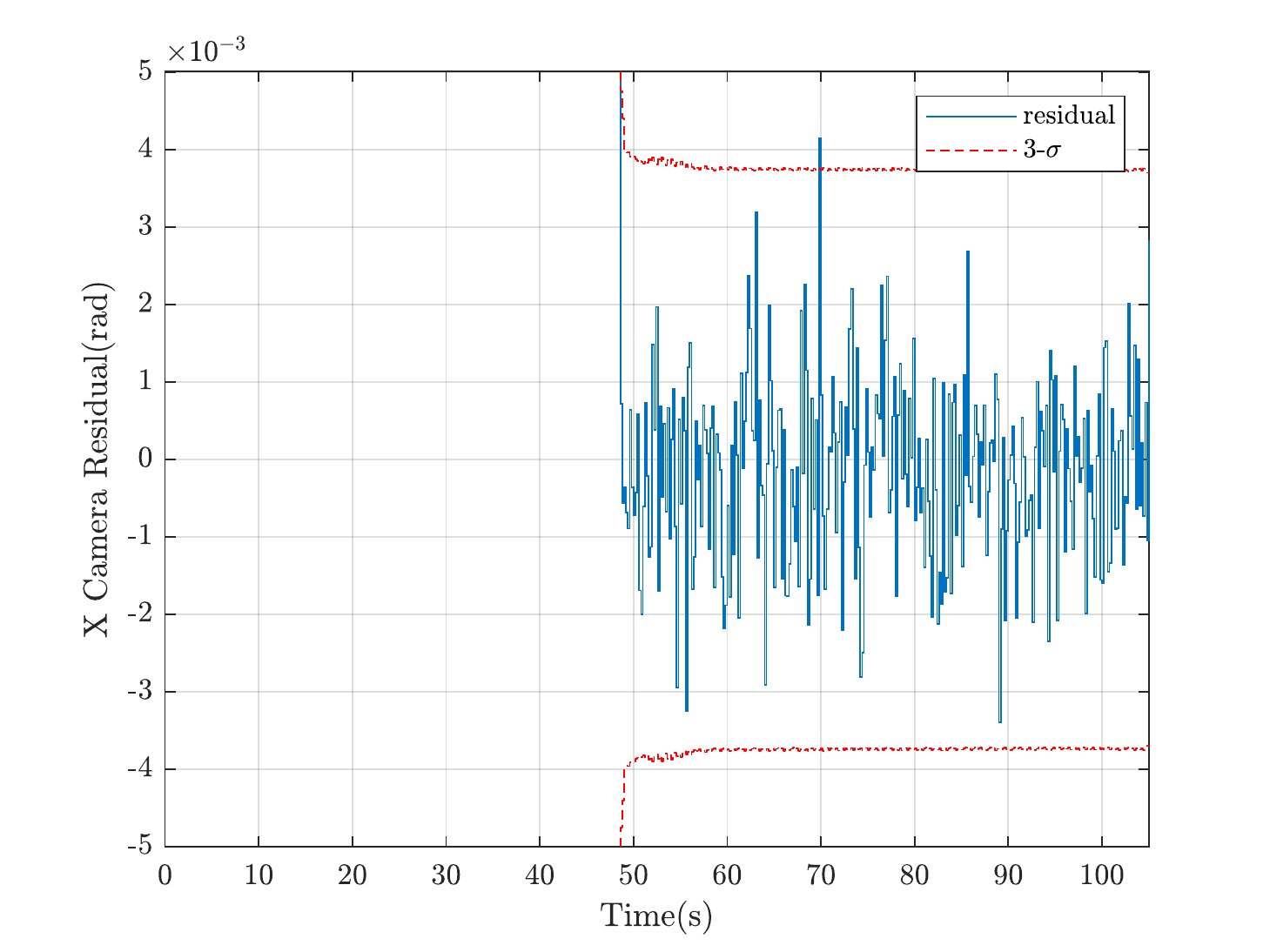}
    \includegraphics[trim={0.9cm 0.2cm 1.4cm 0.4cm},clip,width=0.45\textwidth]{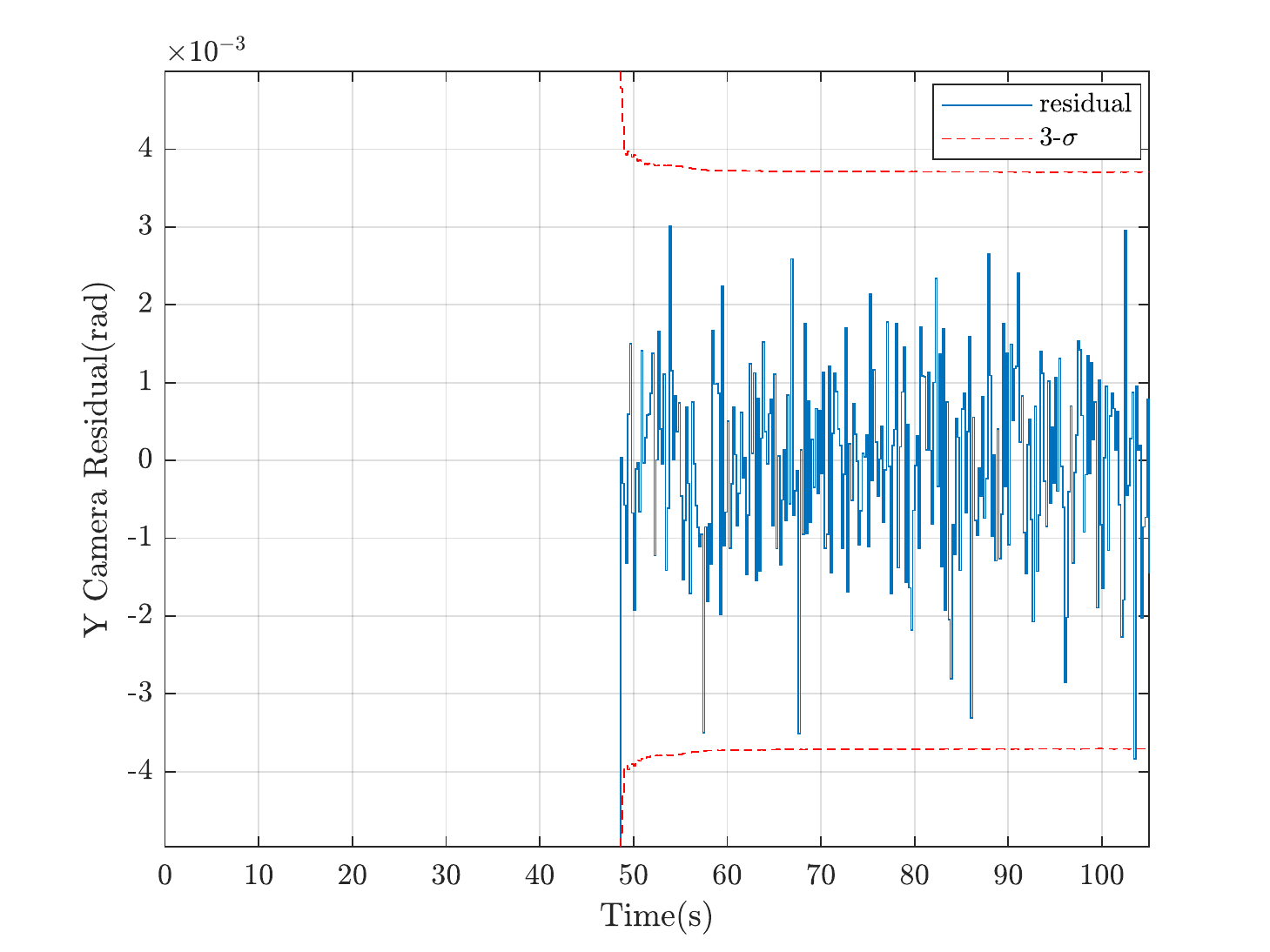}
    \caption{LOS measurement residuals}
    \label{fig:CamResiduals}
\end{figure}

To validate correct implementation of the developed EKF, the measurement residuals are plotted and verified to stay mostly within 3-$\sigma$ bounds and be zero-mean and white.  The GNSS measurement residuals can be seen in Figure \ref{fig:GNSSResiduals}, and the LOS measurement residuals can be see in Figure \ref{fig:CamResiduals}.  Note that the GNSS residuals are only present during the first 48.5 seconds because this is when GNSS measurements are processed, and camera residuals are only present after 48.5 seconds when these measurements are available. 

\begin{figure}
    \centering
    \includegraphics[trim={0.7cm 0.2cm 1.4cm 0.4cm},clip,width=0.45\textwidth]{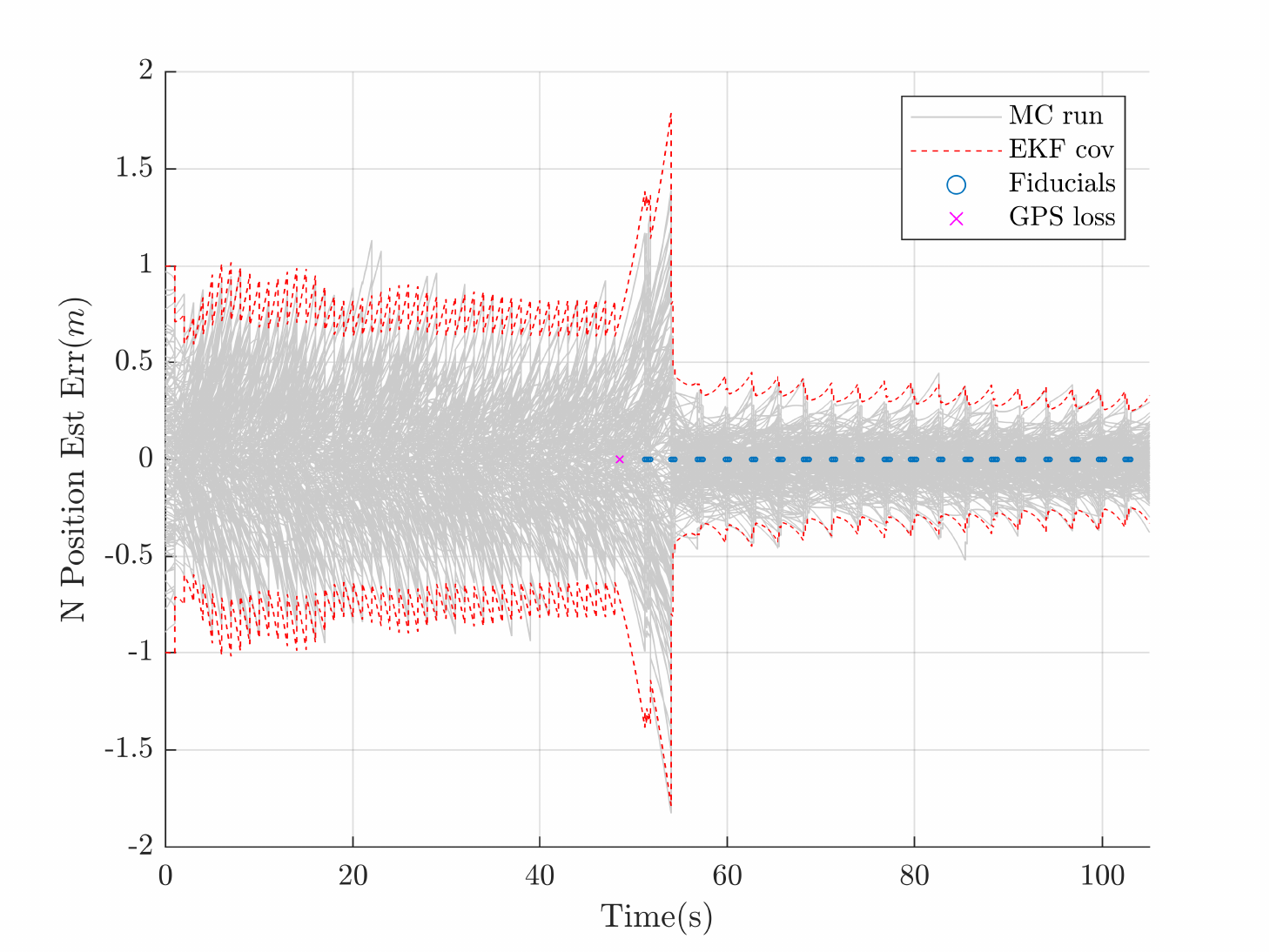}
    \caption{Position estimation error in the north direction}
    \label{fig:EstErrPosN}
\end{figure}

\begin{figure}
    \centering
    \includegraphics[trim={0.7cm 0.2cm 1.4cm 0.4cm},clip,width=0.45\textwidth]{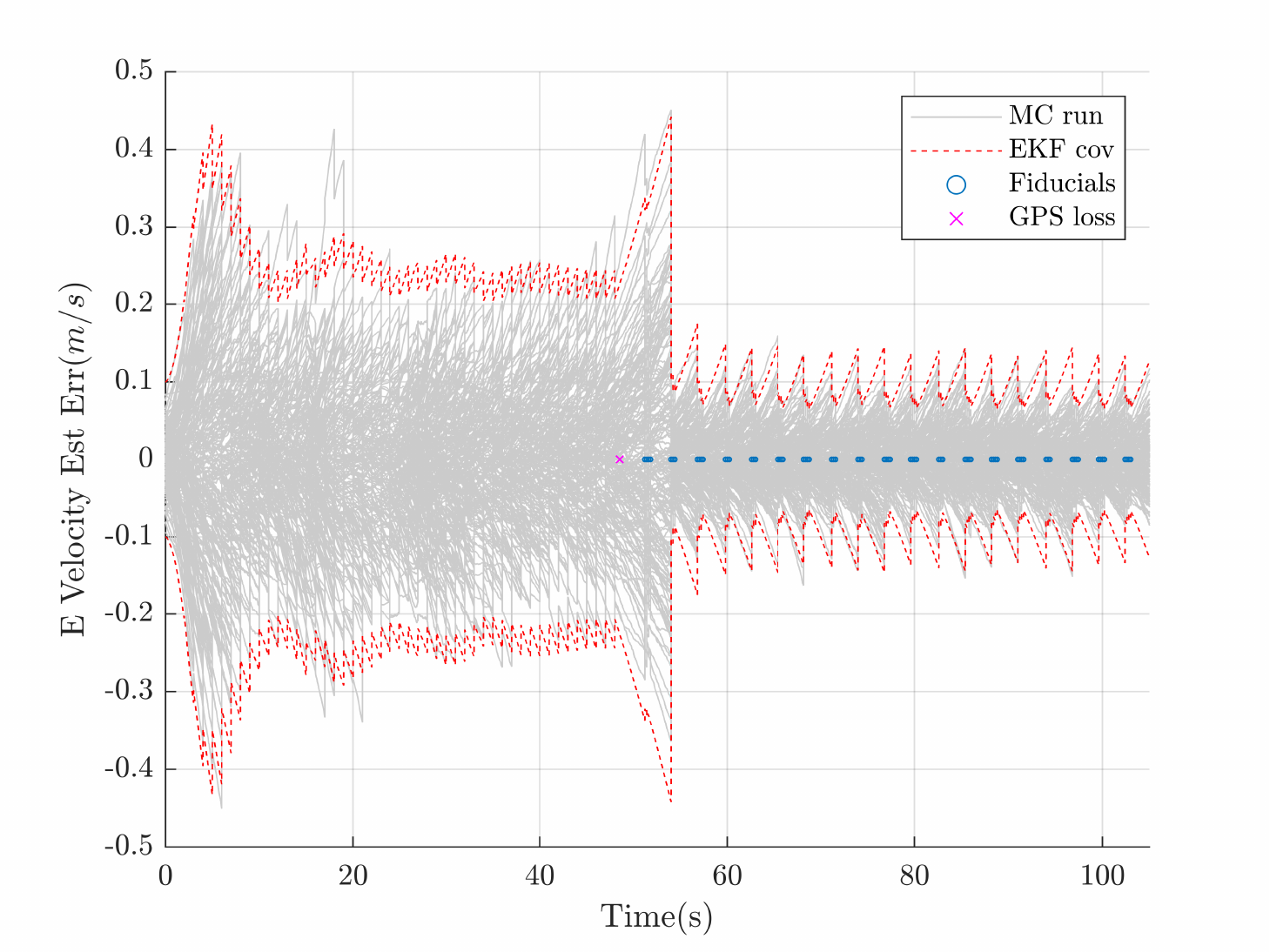}
    \caption{Velocity estimation error in the east direction}
    \label{fig:EstErrVelE}
\end{figure}

\begin{figure}
    \centering
    \includegraphics[trim={0.3cm 0.2cm 1.4cm 0.4cm},clip,width=0.45\textwidth]{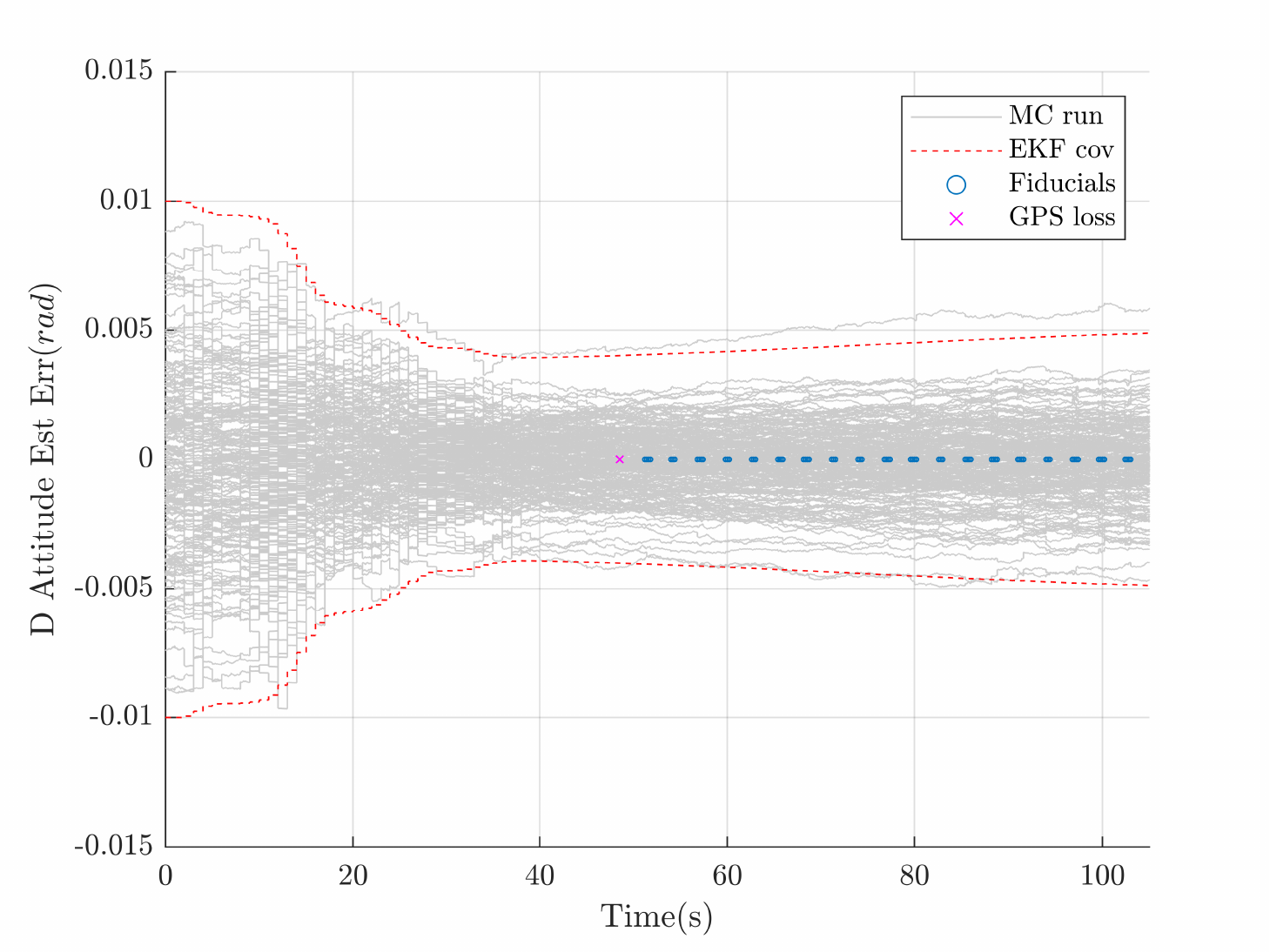}
    \caption{Attitude estimation error in the down direction}
    \label{fig:EstErrAttD}
\end{figure}

The difference between the true states of the UAV and the estimated states is calculated and plotted for 200 Monte Carlo simulations of running the EKF with the given trajectory.  These differences are plotted as the grey hairlines seen in Figures \ref{fig:EstErrPosN} through \ref{fig:EstErrAttD}.  The red lines represent the corresponding 3-$\sigma$ values obtained from the EKF covariance matrix.  The time at which GNSS signals are lost is marked with a red X on the x-axis, and the times at which fiducial measurements are processed are marked with blue dots along the x-axis.  It can be seen that the hairlines are consistent with the 3-$\sigma$ bounds predicted by the EKF covariance matrix.  Figures \ref{fig:EstErrPosN} through \ref{fig:EstErrAttD} show a selection of these plots for one direction of the position, velocity, and attitude.  Similar consistency is observed for all filter states.  It can be seen in the position and velocity plots that once the GNSS signals are lost, the estimation error grows until fiducials begin being processed. The estimation error grows due to the errors in the IMU measurements between the loss of GNSS measurements and availability of LOS measurements. The GNSS and LOS measurements provide absolute position information, causing estimation error to reduce.

\subsection{Sensitivity Analysis}

Now that the EKF has been validated, the covariance matrix can be used to ascertain the sensitivity of the navigation system to varying parameters such as IMU grade, fiducial spacing, LOS measurement processing frequency, and trajectory height above fiducials. The metric used for the sensitivity analyses is the average of the 3-$\sigma$ values for the last third of the simulation from about 70 seconds to 105 seconds, after the EKF covariance has reached a steady-state saw-tooth pattern.

\begin{figure}
    \centering
    \includegraphics[trim={0.8cm 0.2cm 1.1cm 0.5cm},clip,width=0.45\textwidth]{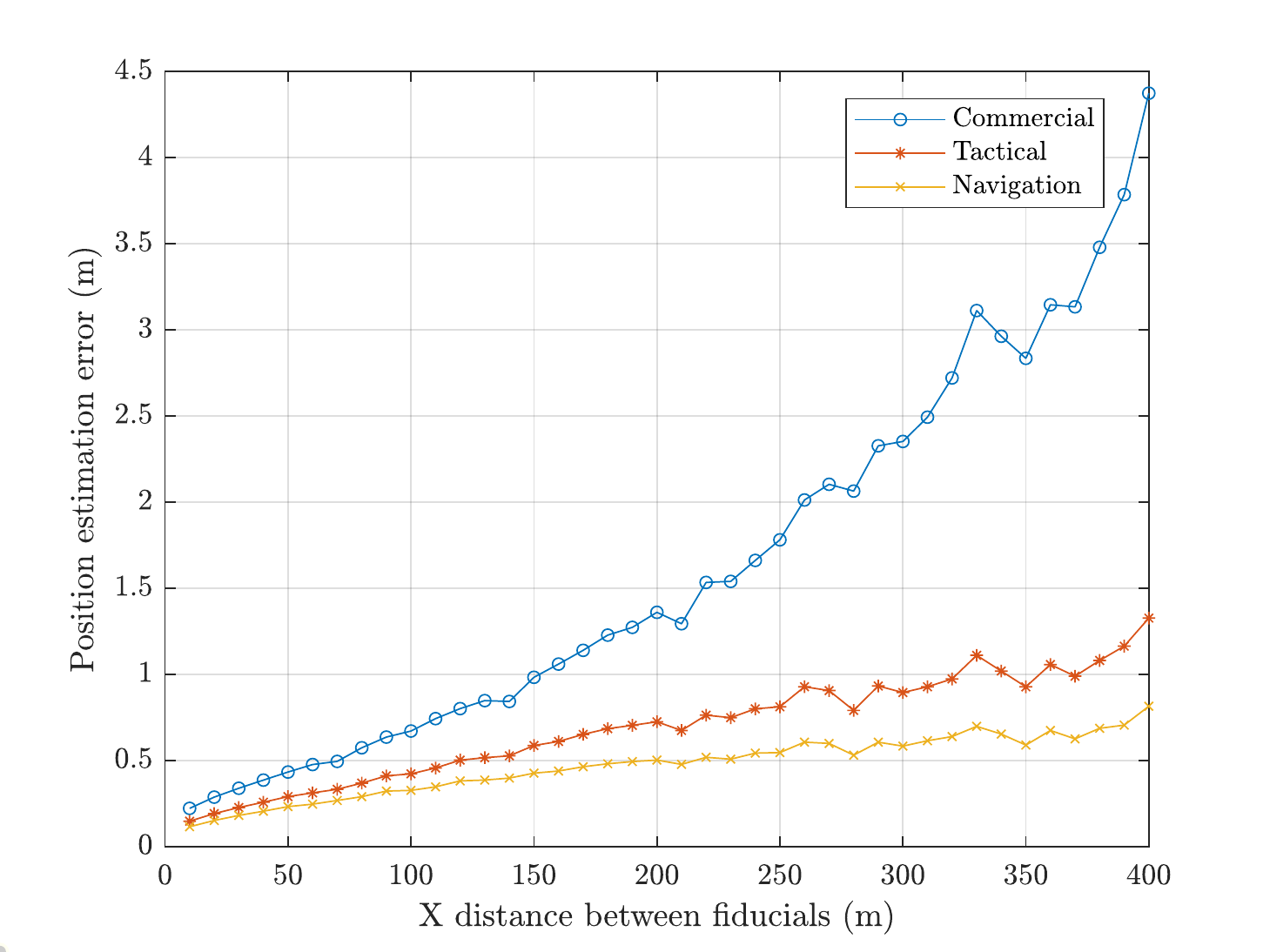}
    \caption{Average 3-$\sigma$ steady state position estimation error for three grades of IMU and various distances between fiducials}
    \label{fig:IMU-FidDensity}
\end{figure}

Figure \ref{fig:IMU-FidDensity} shows the sensitivity of the navigation performance to varying along-track distance between fiducials and varying IMU grade. The y-axis represents the total, or Root-Sum-Squared (RSS), of 3-$\sigma$ position errors, while the fiducial spacing is represented by the x-axis.  The IMU grades are represented by the blue, red, or yellow traces.  

The spacing between fiducials is varied from 10 meters to 400 meters in 10-meter increments.  It can be seen that at the 10-meter spacing, there is only about a 10 centimeter difference between the commercial- and tactical-grade IMUs, while at the 400-meter spacing, there is about a 3.6 meter difference. This shows that when fiducials are spaced closer together, the grade of IMU becomes less and less important.  Given the high cost of navigation-grade IMUs, a potentially more cost effective system might employ high fiducial density to maintain the position errors below desired levels.

\begin{figure}
    \centering
    \includegraphics[trim={0.6cm 0.2cm 1.1cm 0.5cm},clip,width=0.45\textwidth]{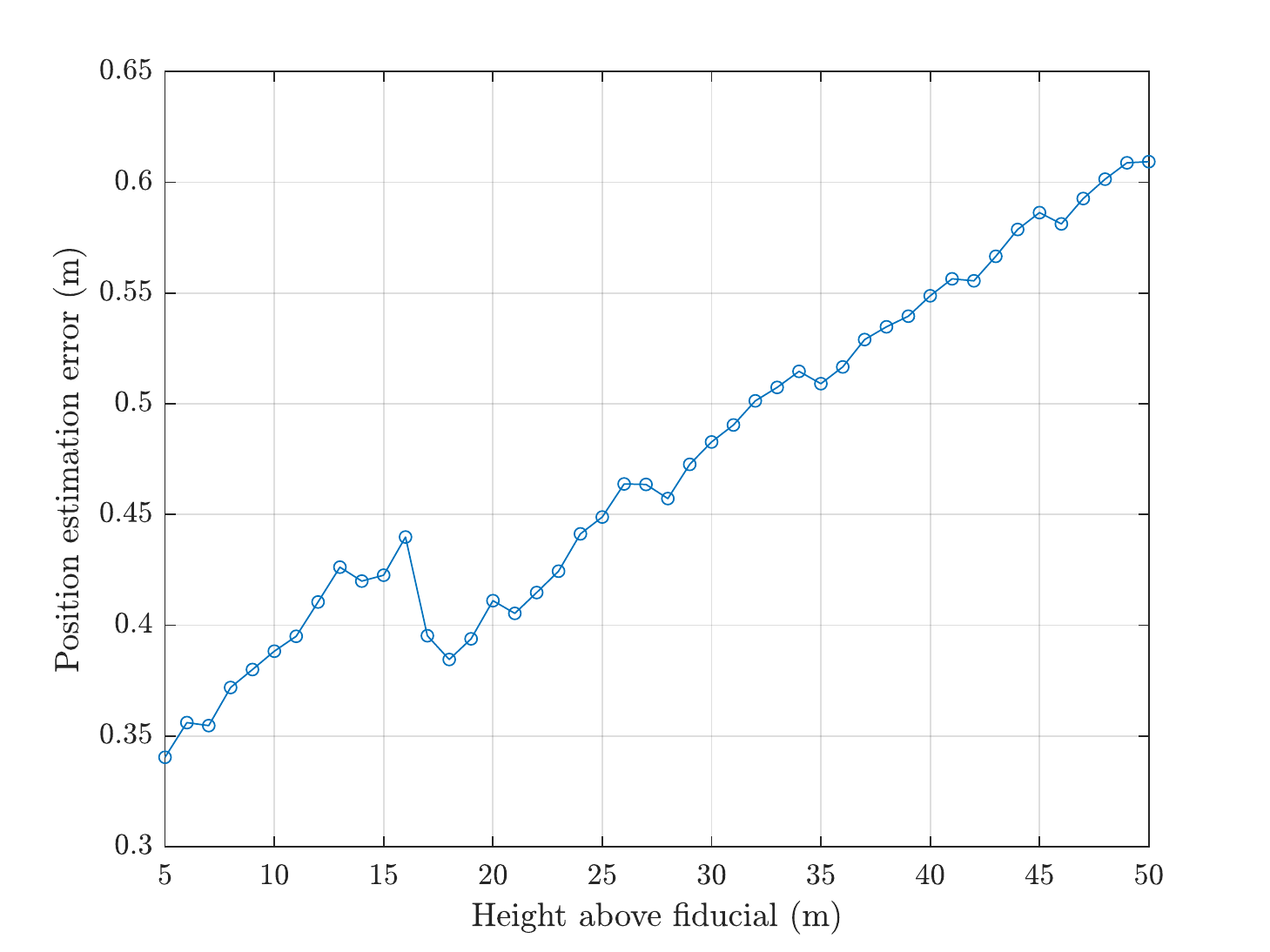}
    \caption{Average 3-$\sigma$ steady state position estimation error as height of the trajectory above the fiducials was varied \revision{with a tactical-grade IMU}}
    \label{fig:HeightAboveFid}
\end{figure}

Figure \ref{fig:HeightAboveFid} shows the effect of varying the height of the trajectory above the fiducials. It can be seen, in general, that the position errors increase with the trajectory height. This is not the case between the 16- and 18-meter spacings.  At 18 meters, the 40 degree FOV combined with the 5 Hz processing frequency causes an additional fiducial to be observed and processed that is not observed and processed at the 16-meter range, thus decreasing the amount of position estimation error.  This shows that ideal fiducial placement needs to be designed with the possible UAV trajectories in mind as well as system parameters such as LOS processing frequency and camera FOV. 

\begin{figure}
    \centering
    \includegraphics[trim={0.8cm 0.2cm 1.1cm 0.5cm},clip,width=.45\textwidth]{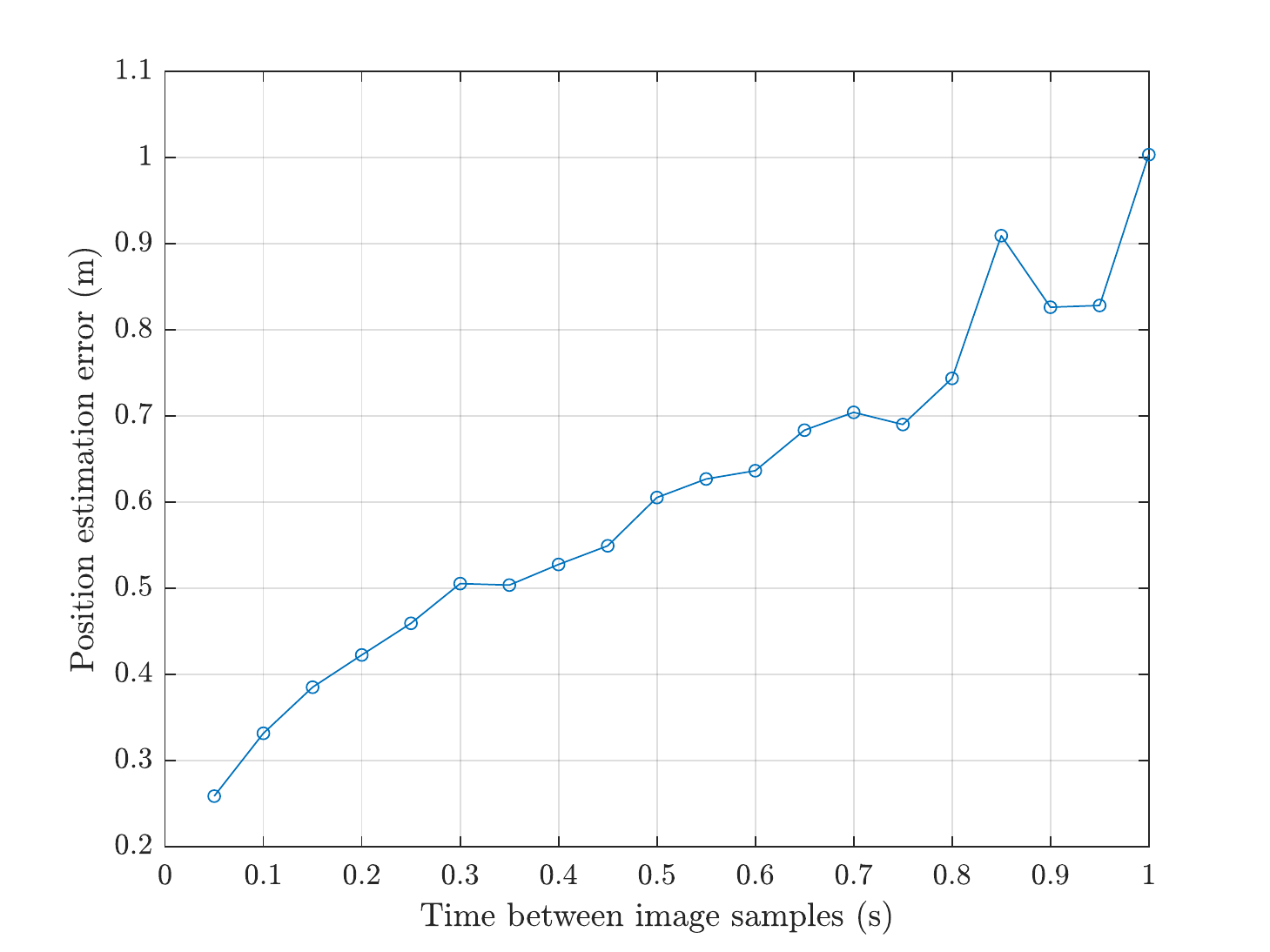}
    \caption{Average 3-$\sigma$ steady state position estimation error as frequency of image processing was varied \revision{with a tactical-grade IMU}}
    \label{fig:ImageFrequency}
\end{figure}

Figure \ref{fig:ImageFrequency} shows that the position errors increase with increased time between LOS measurements.  When LOS measurements are processed at about 20 Hz with all of the other parameters set to the nominal values, the steady-state position estimation error is 0.26 meters.  Processing LOS measurements only once per second results in an increased position error of about one meter.  When time between image captures increases, the performance of the system becomes less predictable.  As images are processed less frequently, it becomes less likely that images will be captured at the precise moments to observe a given fiducial. This makes it less certain that a fiducial will be seen at all as the UAV flies over it.  This demonstrates that the processing frequency of images is an important parameter to consider when determining the best fiducial placement strategy for SDF-aided navigation.

\section{Conclusion}

In this paper, an EKF for the purpose of combining continuous IMU measurements with discrete camera LOS measurements to single-point fiducials was developed and validated.  It was shown that single-point SDF-aided navigation can provide position estimation errors of 0.26 meters or less, depending on fiducial placement, UAV trajectory, and characteristics of the measurement equipment.  This shows that single-point SDF-aided navigation is a viable solution for GNSS-denied navigation of UAVs.

Several trends were observed.  One is that as fiducials are spaced further apart, the position estimation error grows more quickly when commercial-grade IMUs are used instead of tactical- or navigation-grade IMUs. If the fiducials are placed at 10-meter intervals along the UAV's trajectory, there is only about a 10 centimeter difference in the position estimation error obtained using the commercial- and navigation-grade IMUs.  With 400-meter spacing, this difference increases to almost 4 meters.  Even at the 100-meter spacing, the estimation error difference between commercial- and navigation-grade IMUs is only about 34 centimeters.  As navigation-grade IMUs are significantly more expensive than commercial-grade IMUs, a cost-effective strategy for SDF-aided navigation would likely involve more densely-spaced fiducials rather than more expensive IMUs.

Another trend showed that as the UAV's trajectory height above the fiducials increases, the position estimation error also increases.  When the additional height combined with the camera frequency and camera FOV allow for an additional fiducial to be captured and processed, the additional height can reduce estimation error, but in general, the analyses showed that lower altitude UAV trajectories result in reduced position errors.

The final observed trend was that as the time between image samples increases, the position estimation error grows.  This portion of the study also demonstrated that as the time between image captures increased, the trends in estimation error became less predictable and more dependent on the chance that a fiducial was observable when the image was taken. 

Overall, the position estimation error that is acceptable for a given SDF-aided navigation system needs to be considered along with total system cost in order to determine the ideal setup.  This study showed that less expensive means such as more densely-placed fiducials and higher image processing frequencies can provide a significant amount of certainty in position estimation.  Using a more expensive, higher-grade IMU will also help but may be less /revision{desirable from a cost standpoint}.

\revision{Other sensitivities of a fiducial-aided inertial navigation system may be of interest for a particular application. Potential areas of future study include the sensitivity of system performance to camera resolution, fiducial spacing in the cross-track direction, trajectory deviations in the cross-track direction, amongst others. The EKF covariance framework developed and validated in this work provides a means to perform such investigations.}

\section{Declaration of conflicting interests}
The authors declare that there is no conflict of interest.


\end{document}